\documentclass[letterpaper, 10 pt, conference]{ieeeconf}  

\IEEEoverridecommandlockouts                              

\usepackage{graphics} 
\usepackage{color}
\usepackage{graphicx} 
\usepackage{epsfig} 
\usepackage{amsmath, bm} 
\usepackage{amssymb}  
\usepackage{upgreek}
\usepackage{multicol}
\usepackage{multirow}
\usepackage{lipsum}  
\usepackage{breqn}
\usepackage{comment} 
\usepackage{lipsum} 
\usepackage{booktabs}
\PassOptionsToPackage{hyphens}{url}\usepackage{hyperref} 
\usepackage[caption=false]{subfig} 

\usepackage[style=ieee]{biblatex}

\DeclareSourcemap{
  \maps{
    \map{
      \pertype{article}
      \step[fieldset=url, null]
      \step[fieldset=doi, null]
      \step[fieldset=issn, null]
      \step[fieldset=isbn, null]
      \step[fieldset=note, null]
      \step[fieldset=editor, null]
      \step[fieldset=urldate, null]
      \step[fieldset=file, null]
    }
  }
}
\DeclareSourcemap{
  \maps{
    \map{
      \pertype{inproceedings}
      \step[fieldset=url, null]
      \step[fieldset=doi, null]
      \step[fieldset=issn, null]
      \step[fieldset=isbn, null]
      \step[fieldset=note, null]
      \step[fieldset=editor, null]
      \step[fieldset=urldate, null]
      \step[fieldset=file, null]
    }
  }
}
\DeclareSourcemap{
  \maps{
    \map{
      \pertype{incollection}
      \step[fieldset=url, null]
      \step[fieldset=doi, null]
      \step[fieldset=issn, null]
      \step[fieldset=isbn, null]
      \step[fieldset=note, null]
      \step[fieldset=editor, null]
      \step[fieldset=urldate, null]
      \step[fieldset=file, null]
    }
  }
}

\title{\LARGE \bf
Generative Design of NU's \textit{Husky Carbon},\\ A Morpho-Functional, Legged Robot
}

\author{Alireza Ramezani, Pravin Dangol, Eric Sihite, Andrew Lessieur and Peter Kelly $^{1}$
\thanks{$^{1}$The authors are with the Department of Electrical and Computer Engineering, Northeastern University, Boston, MA, USA. SiliconSynapse Laboratory. Emails: \{a.ramezani, dangol.p, e.sihite, lessieur.a, kelly.pe\}@northeastern.edu.}%
}

\begin{document}

\maketitle

\begin{abstract} We report the design of a morpho-functional robot called \textit{Husky Carbon}. Our goal is to integrate two forms of mobility, aerial and quadrupedal legged locomotion, within a single platform. There are prohibitive design restrictions such as tight power budget and payload, which can particularly become important in aerial flights. To address these challenges, we pose a problem called the \textit{Mobility Value of Added Mass (MVAM)} problem. In the MVAM problem, we attempt to allocate mass in our designs such that the energetic performance is affected the least. To solve the MVAM problem, we adopted a generative design approach using Grasshopper's evolutionary solver to synthesize a parametric design space for Husky. Then, this space was searched for the morphologies that could yield a minimized Total Cost Of Transport (TCOT) and payload. This approach revealed that a front heavy quadrupedal robot can achieve a lower TCOT while retaining larger margins on allowable added mass to its design. Based on this framework Husky was built and tested as a front heavy robot.   

\end{abstract}


\section{Introduction}

A bird uses its legs to negotiate ground surfaces in search of food and employs them to jump and fly (for example, when escaping from a predator), at which point the bird's locomotion style and the organs involved become completely different. The legged mobility of the bird involves intermittent interactions and impulsive effects and its aerial mobility entails continuous fluidic-based interactions. The two modes of mobility are fundamentally different in nature and our example has undergone a complex evolutionary journey to obtain its arrays of locomotion specializations such as a light skeleton and efficient, versatile muscles. However, this complex journey has marked important payoffs for the bird. Gains such as ease at dodging predators and foraging led by the bird's multi-modal locomotion apparatus have been key to its formidable resilience in a very dangerous world of larger predators and competitors. 

Robotic biomimicry of animals' multi-modal locomotion can be a significant ordeal. The prohibitive design restrictions such as a tight power budget, limited payload, complex multi-modal actuation and perception, excessive number of active and passive joints involved in each mode, sophisticated control and environment-specific models, just to name a few, have alienated these concepts. It is worth noting that as how multi-modality has secured birds' (and other species) survival in complex environments, similar manifestations in mobile robots \cite{lock_multi-modal_2013} can be rewarding and important, as will be discussed later. 

In this work, we will report our preliminary results in the design and development of a legged robot called \textit{NU's Husky Carbon}. Our objective is to integrate legged and aerial mobility in a single platform. Since total mass and its distribution is important for both legged and aerial locomotion, we will present our approach based on inspecting every possible solution in the design space of the robot, which is led by a generative design algorithm, for a minimized Total Cost Of Transport (TCOT) and payload.   

\begin{figure}[!t]
    \centering
    \includegraphics[width = 0.9 \linewidth]{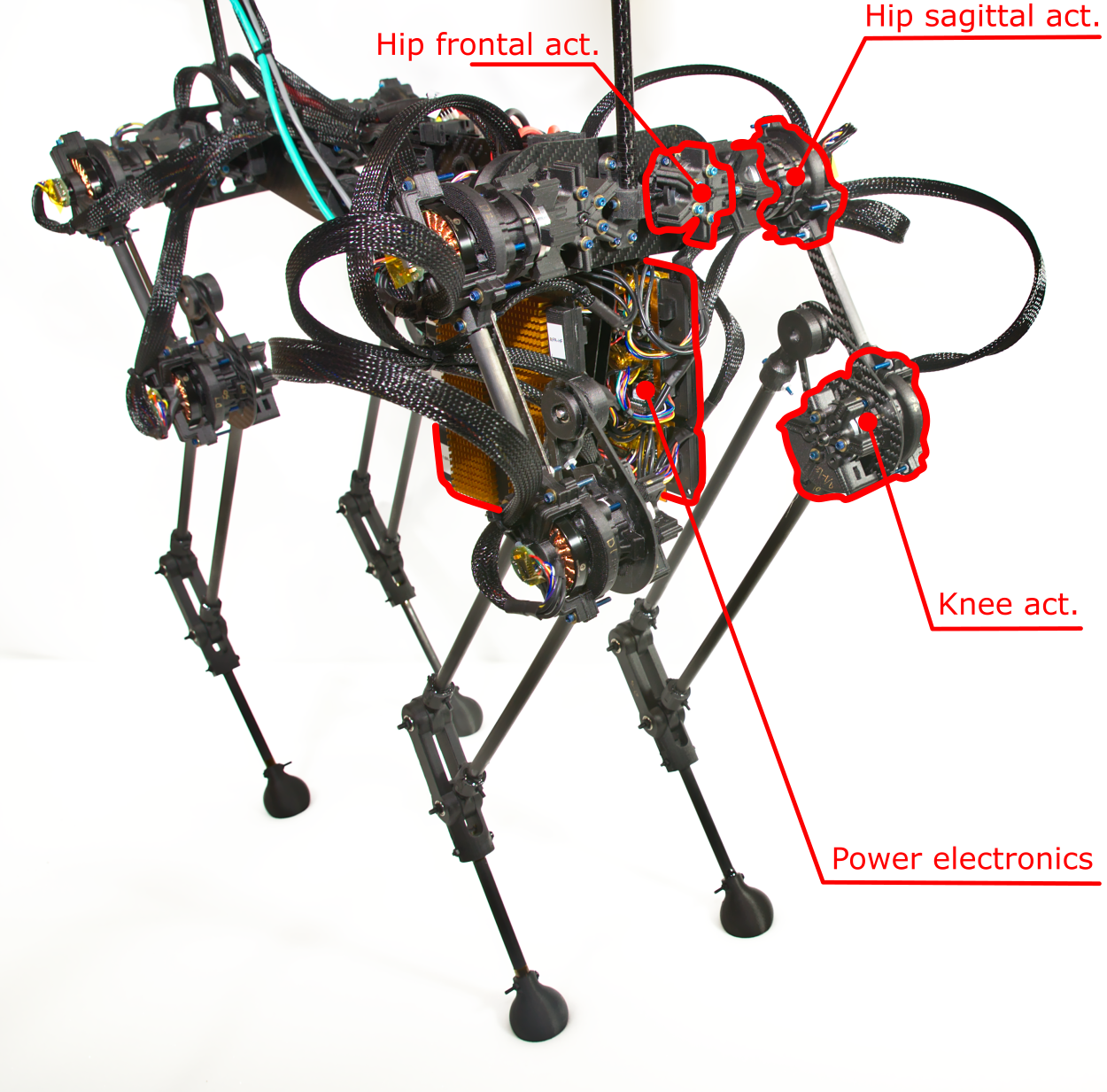}
    \caption{NU's Husky Carbon}
    \label{fig:cover_fig}
\end{figure}

While energetic efficiency of legged locomotion has been extensively studied based on shaping joint trajectory \cite{ramezani_performance_2014, dangol2020towards,dangol2020performance,liang2021roughterrain,de2020thruster}, actuator design \cite{wensing_proprioceptive_2017,seok_design_2015} and compliance \cite{kashiri_overview_2018,hutter_starleth:_2012}, there has been little to no attempt to connect it to robot morphology. For instance, efficiency was a key factor in the MIT Cheetah robot's trotting gaits with jerky and interrupted joint movements under large Ground Reaction Forces (GRFs) \cite{park2017high,hyun_high_2014}. Efficiency was achieved based on electric actuator design and selection of predefined trotting gaits. Since commercial actuators with high ratio gearboxes are often tightly designed based on more continuous and controlled behaviors, the Cheetah possessed backdrivable and brushless actuators. 

In ANYmal, the emphasis has been on endurance and with its bulky structure, so natural and fast running patterns showcased by MIT's Cheetah are not feasible on this platform \cite{hutter_anymal_2016}. ANYmal's design characteristics are high torque joint actuation through the combination of electric motors and harmonic drives with rotational springs, precision joint movement measurement, and a rigid, bulky leg structure that accommodate high mechanical bandwidth for high bandwidth impedance control. With these characteristics, robot morphology and payload have not been important in the design of ETH's ANYmal. 

Before introducing NU's Husky platform, explaining the challenges and outlining our design philosophy, we will answer the following question \textit{''What could be the merits of a legged-aerial platform?''}   

\subsection{Lessons from Past Disasters}

The major motivation is to create a platform that possesses the fast mobility (at high altitudes) of an aerial system and the safe, agile and efficient mobility of a quadrupedal system in unstructured spaces. 

In Search And Rescue (SAR) operations and in the aftermath of unique incidents such as flooding, one event may accompany another disaster. A hurricane may produce flooding as well as wind damage, or a landslide may dam a river and create a flood. In these scenarios, mono-modal mobile systems can easily fail. For instance, Unmanned Aerial Systems (UAS) can deliver important strategic situational awareness involving aerial survey and reconnaissance, which can be key in locating victims quickly, through scans of the area with their suite of sensors. However, airborne structural inspection of buildings in harsh atmospheric conditions is challenging if not impossible. In addition, aerial mobility is not feasible inside collapsed buildings. To inspect inside these structures, legged mobility in form of crawling and walking is superior to aerial or wheeled mobilities \cite{alexander_principles_2013}.

Another motivation is the safety of mobility. The absence of safe UAS was costly in the aftermath of Hurricane Katrina in 2005. In this incident, which set the stage for drone deployments, regulations promulgated by the Federal Aviation Administration (FAA) posed severe limitations on drone operations in inflicted regions. According to the FAA, extreme care must be given to and when flying near people because operators tend to lose perception of depth and may get far too close to objects and people. 

\begin{figure*}[!t]
    \centering
    \includegraphics[width = 0.9 \linewidth]{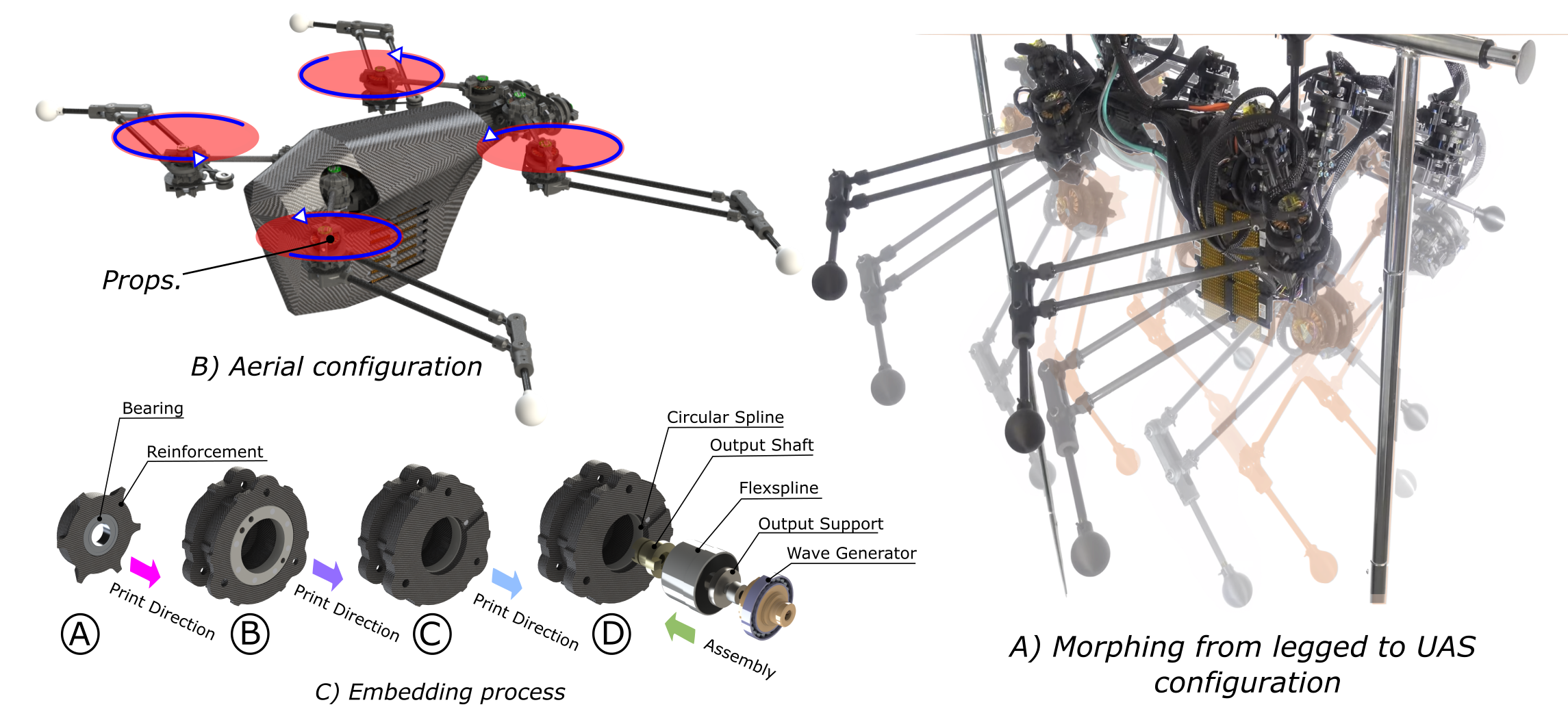}
    \caption{Illustration of A) morphing from legged to UAS configuration which is possible because of the large range of motion in hip frontal actuators; B) the concept UAS configuration; C) the embedding process of actuators' components within the body structure.}
    \label{fig:husky_overview}
\end{figure*}

Quadrotors and other rotorcraft require safe and collision-free task spaces for operation since they are not collision-tolerant and have rigid body structures. The incorporation of soft and flexible materials or rotor shields into the design of such systems has become common in recent years, but the demands for aerodynamic efficiency prohibit the use of propellers made of extremely flexible materials. Consequently, these systems are not only an impractical solution for cluttered environments such as inside buildings, but also are inherently unsafe for victims, since the rotating propellers can inflict severe bodily injuries.  

\section{NU's Morpho-functional Platform: Husky Carbon}

\subsection{Overview}

Husky Carbon, shown in Fig.~\ref{fig:husky_overview}, when standing as a quadrupedal robot, is 2.5 ft (0.8 m) tall. The robot is about 1 ft (0.3 m) wide. It is fabricated from reinforced thermoplastic materials through additive manufacturing and weighs 9.5 lb (4.3 kg). It hosts on-board power electronics and, currently, it operates using an external power supply. The current prototype lacks exteroceptive sensors such as camera and LiDAR. The robot is constructed of two pairs of identical legs in the form of parallelogram mechanisms. Each with three degrees-of-freedom (DOFs), the legs are fixated to Husky's torso by a one-DOF revolute joint with a large range of motion. As a result, the legs can be located sideways as shown in Fig.~\ref{fig:husky_overview}. This configuration allows facing the knee actuators upwards for propulsion purpose. A clutch mechanism will disengage the knee actuator from the lower limb before the actuator runs a propeller. More technical details about the knee actuator and the clutch design will be reported in future publications that address the aerial mobility of Husky Carbon.

The robot possesses a total of twelve actuator. Each custom-made actuator has a pancake brushless-DC motor winding from T-motor with a KV equal to 400. Harmonic drive component sets (flexspline, circular spline and wave generator) with gear ratios: 30, 50 and 100 for knee, hip sagittal and hip frontal, respectively, and hall-effect-based incremental encoders are embedded inside the 3D-printed structure of the robot. The embedding process has minimized the use of metal housings and fasteners and has yielded a lighter structure. The current integration has made Husky self-sustained for stable, blind walking and trotting gaits. We note that currently the platform relies on an off-board body orientation sensing system which soon will be replaced with an on-board inertial measurement unit with negligible weight. 

Next, we will briefly describe the challenges faced in the design and control of Husky. We will layout our approach and explain how we obtained the current design by drawing a connection between morphology and efficiency using MATLAB and generative design tools such as Grasshopper, which is a visual programming language within Rhinoceros 3D computer-aided design application. We note that our control design paradigm have been submitted in a separate work to the similar conference. 
 
\begin{figure*}[!t]
    \centering
    \includegraphics[width = 0.9 \linewidth]{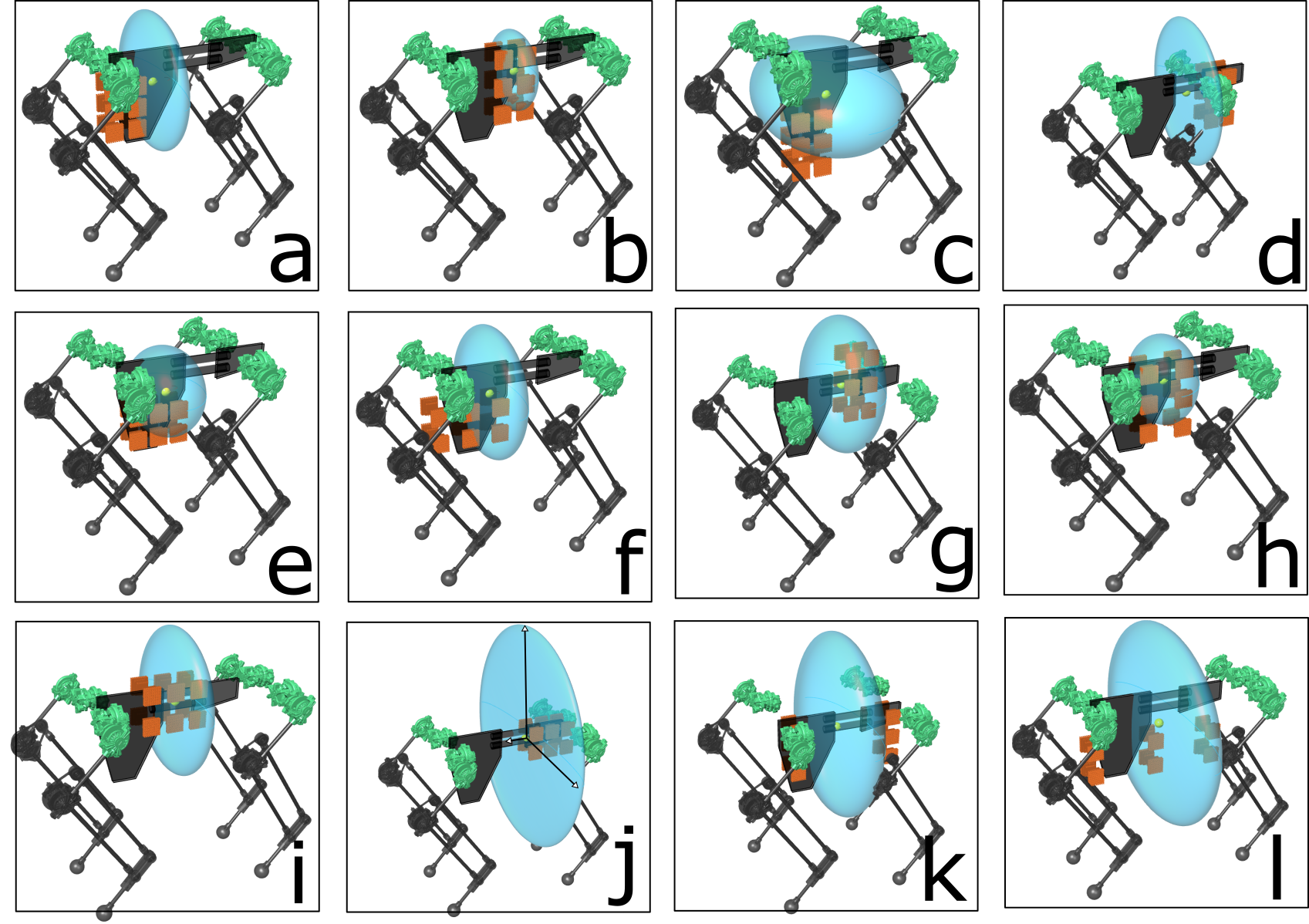}
    \caption{Illustration of a part of the design space led by a computer-aided, generative design algorithm. A parametric model and an evolutionary solver from Grasshopper describe the design space which is linked to a parametric physics model in MATLAB. Various morphologies based on components' locations, size and shapes are searched for a minimized TCOT and payload.}
    \label{fig:gen_design}
\end{figure*}

\subsection{Design Challenges}

In the design of Husky, we had to combat prohibitive design restrictions such as payload and actuation power. By looking at legged locomotion as the translation of the COM in the environment led by body self-manipulation it is easy to notice that added mass from structures, actuators, electronics, etc., can yield added required actuation torques at the joints and have to be compensated by stronger and larger actuators which in turn add extra mass to the system. So, the resulting loop of added mass for the added required joint torque can lead to bulky systems. This issue becomes very important when aerial mobility is considered as, for example, the application of metal parts or bulky actuators that are widely used in legged robots are not anymore possible. That said, the use of lighter non-metallic structures, unreinforced body frameworks, smaller actuators, etc., can manifest itself in destructive ways such as unplanned compliance or lack of actuation power in the system. 


Compliance can yield control design challenges both for aerial and legged mobility. In general, whole body control in legged systems is challenging by itself and unplanned compliance can lead to extra  \cite{dario_bellicoso_perception-less_2016,farshidian_robust_2017}. That said, compliance is not a negative property by itself and is the defining characteristics of biological locomotion systems \cite{alexander_principles_2013}. This is the reason legged community has adopted series elastic actuators \cite{roy_hybrid_2013,hutter_efficient_2013}. 

The inherent compliance in Husky's body has led to a few issues which have been addressed with the help of closed-loop feedback. While Husky's compliant legs can potentially accommodate locomotion on grounds that feature unknown irregularities, body flexibility in Husky can prohibit precise kinematic planning for perceived ground changes and can lead to instantaneous body sagging each time feet touch-downs occur. Other issues including degraded mechanical bandwidth as it is important to the projection of joint torques to body forces, impedance control, or foot placement has been observed in Husky. For instance, the leg compliance can introduce oscillations of large amplitude and can lower the natural frequency of the tracking controller meant to position the COM at desired locations with respect to the contact points \cite{townsend_mechanical_1989,fankhauser_robust_2018,hutter_efficient_2013,winkler_gait_2018,dario_bellicoso_perception-less_2016}.


Quadrupedal robots are often designed in such a way that they have wide support polygons which allow trivial, quasi-static locomotion \cite{winkler_gait_2018,mastalli_-line_2015,farshidian_robust_2017,farshidian_robust_2017}. Husky possesses a very small stability margin based on a crude definition \cite{kimura_adaptive_2007} which considers the shortest distance from the robot's projected COM to the edges of its support polygon constructed by the contact points. This small margin is led because of two reasons. First, the small stability margin is caused partly because of a low cross-section torso -- i.e., the hips are located very close to each other in the frontal plane -- which is important to reduce induced drag forces for aerial mobility. Second, it is partly caused by the inherent compliance in the system which can cause fall-overs even when the robot has four contact points. Due to the robot body flexibility COM can shift towards the edges of the support polygon very easily if required corrections are not made quickly through foot placement or active control of the joints.     

Small stability margins in Husky has led to other challenges such as gait design issues that generally are expected in bipedal systems and not quadrupedal robots \cite{apgar_fast_2018}. In Husky, the stability margin and gait cycle periods are adversely related. When the gait cycle time is larger than approximately one third of a second, which is equivalent to a 3-Hz gait cycle, even when the robot retains three contact points with the ground surface the projected position of the robot COM could reach to the boundaries of the support polygon. If new foot placements are not involved the robot's stability margin will reduce further by the sagging effects led by the compliance in the robot. This has posed severe gait design challenges for Husky.

\subsection{Mobility Value of Added Mass (MVAM) Problem}

Added mass can affect various aspects of mobility in a mobile robot, e.g., in legged systems, it can affect impact dynamics, energetic efficiency, selection of actuators, etc. It is nearly impossible to consider every aspect of mobility when dealing with a mass allocation problem. In this paper, we will only consider the connection between morphology and energetic efficiency of legged mobility which has direct implications in the design of NU's Husky. We call this problem \textit{Mobility Value of Added Mass (MVAM).} While aerial mobility is another important objective in this project, its technical details are beyond the scope of this paper and the MVAM problem only addresses payload restrictions in relation to it. 

Inspection of energetic efficiency of legged robots is not a new topic and has been extensively studied and various key players such as joint movement, actuator design and compliance have been examined. However, there has been little to no attempt to connect efficiency to robot morphology. In pursuit of a multi-modal design that can realize legged and aerial mobility actuation and payload pose formidable challenges. We are interested to know how Husky's morphology can affect its Total Cost Of Transport (TCOT). The interesting research question that has been tried to answer in the MVAM problem is that how the search for a morphology with minimum TCOT and payload can be mathematically formulated. 

To do this, we drew a connection between generative design algorithms such as those available in Rhino and physics modeling tools from MATLAB. Using Grasshopper, which is a plugin for the Rhino 6 CAD software, we were able to create a parametric model of the entire robot. The parameters including the location of metal parts (e.g., components of harmonic drives) and electronics within body structures, the size of housings, body frame, connecting rods, etc., were made accessible to a Galapagos, an evolutionary solver within Grasshopper, which was allowed to automatically find the overall design space based on any other components that were included in the model. The resulting design space can reflect important aspects including geometry, inertial properties, structural strength and kinematics of the physical model realistically. Tweaks to the geometry of the robot can also be made by the solver. Then, we linked this space to the parametric constrained dynamical model of Husky in MATLAB. 

\begin{figure}[!h]
    \centering
    \includegraphics[width = .7 \linewidth]{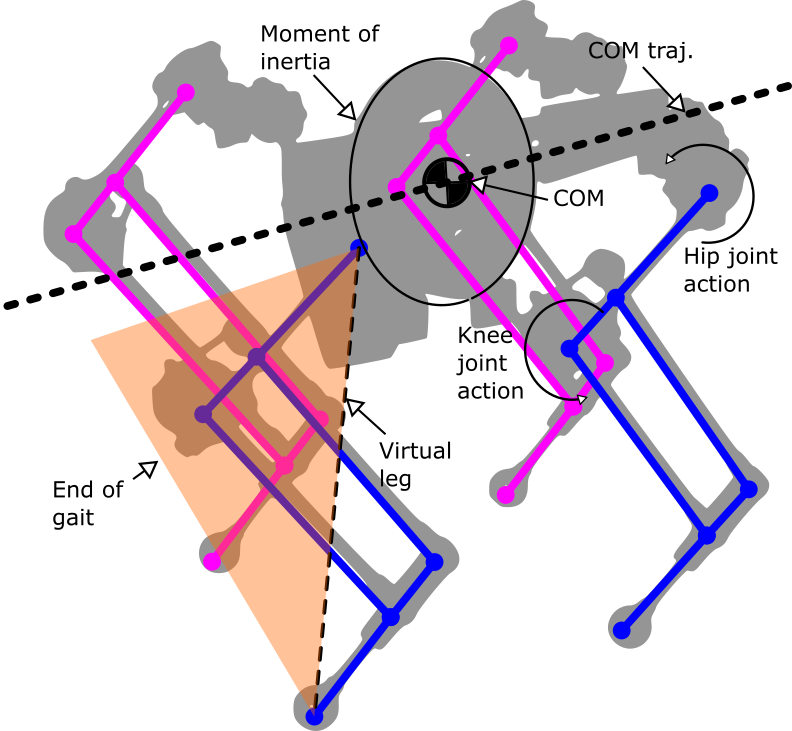}
    \caption{Illustration of a constraint (dashed black line) enforced on Husky's COM for forward walking. This way all morphologies ($m_i\in \mathcal{M}$) exhibit a similar gait pattern.}
    \label{fig:chrom}
\end{figure}

Since finding a closed-form solution of TCOT in terms of morphology (i.e., components location, shape and size) is not feasible we considered the numerical model of Husky in MATLAB. In this model, inertial properties including the COM position and principal mass moment of inertia are introduced as decision parameters which are set by a finite-state nonlinear optimizer in MATLAB. The model is constrained in its sagittal and frontal planes of walking (shown in Fig.~\ref{fig:chrom}). No motion in the frontal plane of locomotion is allowed. In the sagittal plane, all joints movements are forced to follow feasible, predefined trajectories. These predefined trajectories described forward walking gaits in our robot. Contrary to trajectory optimization that assumes a fixed robot morphology, here, fixed gaits and trajectories are assumed and instead the robot morphology is changed. 

The overall search problem can be written for $i=1,\dots n$ , where $n$ is the number of morphologies involved, as following:

\[
\min _{m_i\in\mathcal{M}} \int_{0}^{t_{f}} TCOT(m_i, t, x(t), u(t))
\]
subject to:
\[
\begin{aligned}
\dot{x}(t) &=f_{i}(m_i, t, x(t), u(t)) \\
y &=h\left(x(t)\right) 
\end{aligned}
\]

\noindent where $m_i\in \mathcal{M}$ is the space of all possible designs and morphologies. Gait time period, state and joint torque vectors are denoted by $t_f$, $x(t)$ and $u(t)$, respectively. The dynamics for each morphology are denoted by $f_i(.)$ and $h(x(t))$ denotes the constraints enforced to the models to ensure all follow a similar gait pattern (i.e., forward walking). 
 
\subsection{Additive Manufacturing and Embedding Components} 

Consumables applied in the metal housing in standard actuators and other supporting components are considered as the bulk of mass in legged robots. Actuators compensate major external loads and any design flaw would lead to the concentration of forces internally and consequently yield failures. As a result, designers aim at higher safety factors and often achieve extra hardware security through massive metal housing and support parts. 

To avoid the problem, the fabrication of Husky took place based on embedding electronics and mechanical components precisely at the locations found by MVAM problem. We applied a MarkForged 3D printer in this section. For instance, the steps involved at embedding a harmonic drive's component sets are shown in Fig.~\ref{fig:husky_overview}. In this process, circular splines were embedded inside the actuator housings and other components including wave generators and flexsplines were embedded at the subsequent steps. Embedding components within 3D printed structures has been reported before and it is not a novel approach, however, linking this approach to a physics-based generative design method (MVAM problem) and the use of the results directly to create a legged robot with many parts and components is unprecedented and has not reported before.

The benefits of this approach are as following. First, a considerable payload reduction can be achieved for a feasible and working platform. Second, system components can be precisely located at predetermined locations dictated by the MVAM solutions which are aimed at enhancing efficiency and reduce payload. Last, model and robot can be matched better which is important for model-based control design. Therefore, morphology and its role in control can be taken into account. We did not specifically study the role of morphology in Husky's control. However, we incorporated classical actuator models in our MVAM problem to determine actuation requirements. The findings were used later to select DC motor windings and harmonic drives. 

To minimize added mass over yield strength ratio, we applied a simple approach. To increase the stiffness of light 3D printed parts we applied a method which has roots in classical beam theory. One can increase stiffness of beam by increasing the moment of inertia of the beam section or the modulus elasticity of the material. In our case it was more feasible, from a design perspective, to increase the moment of inertia because we wanted to avoid using heavy metals with a considerably higher modulus. By sandwiching 3D-printed thermoplastic material between layers of continuous carbon fiber weight was reduced while stiffness was greatly increased. This was achieved using MarkForged slicer application called Eiger.

\section{Results and Discussions}

\begin{figure*}
    \centering
    \includegraphics[width = 1.0 \linewidth]{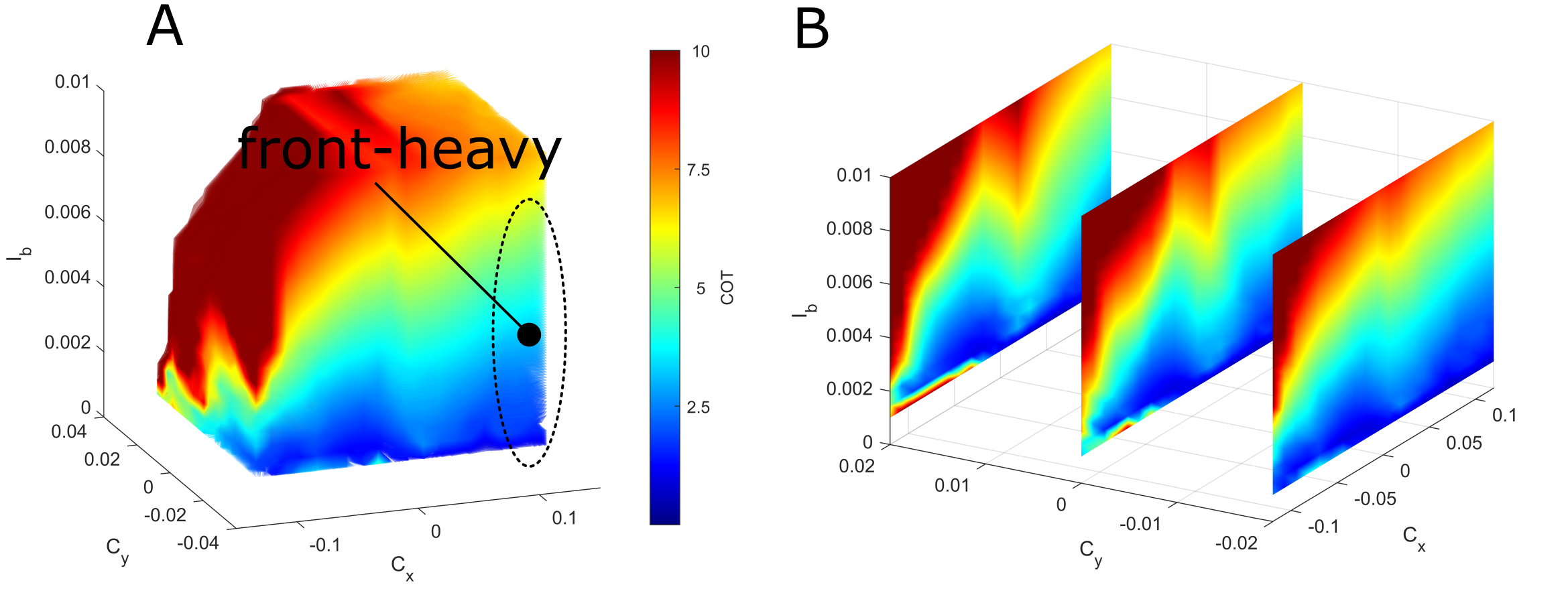}
    \caption{A) Illustrates TCOT for the entire, generated design space represented in terms of the density and color of the data points. B) Cross-sections at the locations with fixed $C_y$ depict how TCOT depends on $C_x$ and $I_b$ in front and back heavy morphologies.}
    \label{fig:tcot}
\end{figure*}

\begin{figure}
    \centering
    \includegraphics[width = 0.8\linewidth]{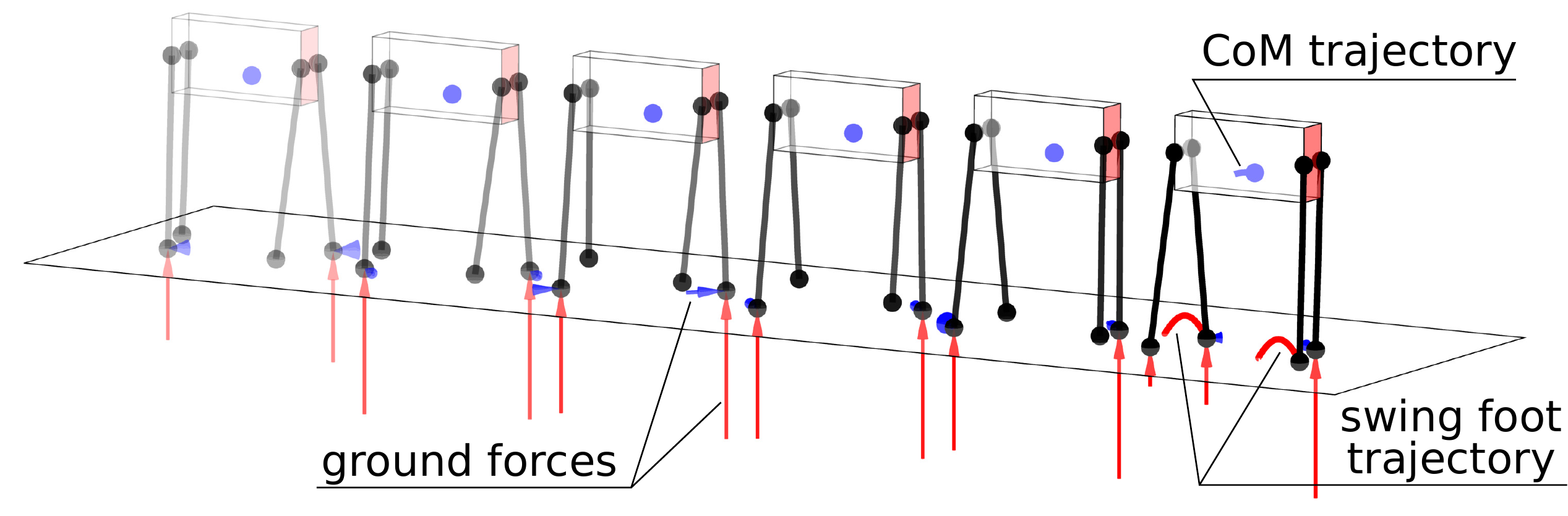}
    \caption{Stick-diagrams illustrate closed-loop forward walking of a front heavy morphology.}
    \label{fig:stickdiagram}
\end{figure}

\begin{figure}
    \centering
    \includegraphics[width = 0.9\linewidth]{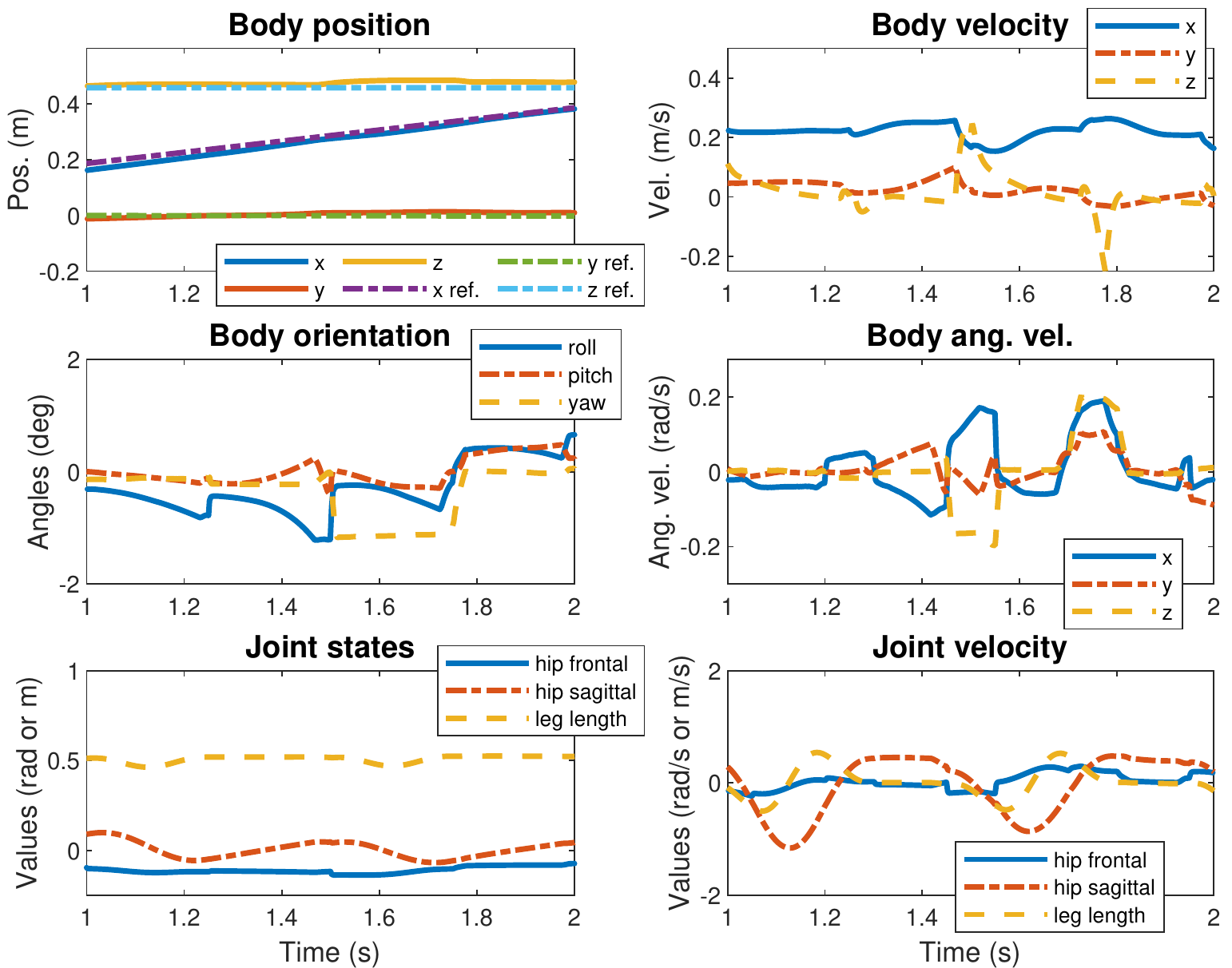}
    \caption{Simulated state-space variables as the model walks forward at the rate of 0.2 m/s and a gait period of 0.25 s. TCOT is 0.2.}
    \label{fig:simulation}
\end{figure}

In Fig.~\ref{fig:tcot}-(A), TCOT values are computed for all of the morphologies in the design space and are represented by the color and intensity of the data points. The low TCOT regions are indicated by the blue and light blue areas. For ease in discussing the results, we will categorize the design space -- some examples from the design space are shown in Fig.~\ref{fig:gen_design} -- led by the generative algorithm into front or back heavy and top or bottom heavy solutions which refer to $C_x$ and $C_y$ positions relative to the geometric center of the robot. The mass moment of inertia at the COM in the sagittal plane is denoted by $I_b$. To show how the design space is characterized by TCOT, we took cross-sections of our data set, shown in Fig.~\ref{fig:tcot}-(B). While front or back heavy solutions are significantly different in terms of TCOT, top and bottom heavy morphologies have little to no effect on the efficiency metric. In Fig.~\ref{fig:tcot}-(B), each cross-section in the $C_y$ direction show generally the same trend. 

In forward walking, we have found that hosting larger added mass in a front heavy morphology is less costly. In other words, one can see (Fig.~\ref{fig:tcot}-A) for a front heavy morphology, lower TCOT can be achieved while larger margins -- or upper bounds -- on the mass moment of inertia in the sagittal plane of locomotion are retained. This finding led us to select a front heavy morphology. We note that since pre-defined, forward walking gaits are considered in this design process the results and conclusions can be biased. That said, we made the design decision to keep the scope of this problem limited. 

Preliminary untethered trotting results were obtained using the front heavy morphology where OptiTrack was used to estimate the position and orientation of the robot. An external power supply was used to energize the robot. Current experimental results may not be able to answer questions in regard to Husky's performance. As a result, more tests and experiments are required to reveal more information about the platform's TCOT. Therefore, we predicted the TCOT in simulation. Simulation results for multiple steps using the unconstrained, front heavy model of Husky were obtained. In this simulation, the lower legs are modeled to be massless and the foot-end positions are defined using leg length and hip angles. Figures~\ref{fig:stickdiagram} and \ref{fig:simulation} illustrate the simulation results for forward walking at a rate of 0.2 m/s and gait period of 0.25 s. This gait is stabilised using a whole body control design approach and has a TCOT equal to 0.2.

\section{Concluding Remarks}

This work reported the design of a morpho-functional robot called Husky Carbon, which is aimed to host two forms of mobility, aerial and quadrupedal legged locomotion. To address challenges such as a tight power budget and payload, a problem called the MVAM problem was posed. First, a generative design approach using Grasshopper's evolutionary solver to synthesize a parametric design space for Husky was employed. Then, this space was searched for the morphologies that could yield a minimized TCOT and payload. Based on this framework Husky was built and tested as a robot with a front heavy morphology.

\printbibliography

\end{document}